\documentclass[10pt,twocolumn,letterpaper]{article}

\usepackage[pagenumbers]{cvpr} 

\usepackage[accsupp]{axessibility}  
\usepackage{graphicx}
\usepackage{amsmath}
\usepackage{amssymb}
\usepackage{booktabs}
\usepackage{algorithm}
\usepackage{algorithmic}
\usepackage{colortbl}
\usepackage{arydshln}
\usepackage{multirow}
\usepackage{multicol}
\usepackage{listings}
\usepackage{xcolor}
\definecolor{codeblue}{rgb}{0.25,0.5,0.25}
\usepackage[pagebackref,breaklinks,colorlinks]{hyperref}

\usepackage[capitalize]{cleveref}
\crefname{section}{Sec.}{Secs.}
\Crefname{section}{Section}{Sections}
\Crefname{table}{Table}{Tables}
\crefname{table}{Tab.}{Tabs.}

\def\cvprPaperID{8146} 
\def\confName{CVPR}
\def\confYear{2022}

\begin{document}
	
\title{QS-Attn: Query-Selected Attention for Contrastive Learning in I2I Translation}

\author{$\text{Xueqi Hu}^1$, $\text{Xinyue Zhou}^1$, $\text{Qiusheng Huang}^1$, $\text{Zhengyi Shi}^1$, $\text{Li Sun}^{1,2}$\footnotemark[1], $\text{Qingli Li}^1$\\
$^1$Shanghai Key Laboratory of Multidimensional Information Processing, \\
$^2$Key Laboratory of Advanced Theory and Application in Statistics and Data Science,\\ 
East China Normal University, Shanghai, China}
\maketitle

\begin{abstract}
Unpaired image-to-image (I2I) translation often requires to maximize the mutual information between the source and the translated images across different domains, which is critical for the generator to keep the source content and prevent it from unnecessary modifications. The self-supervised contrastive learning has already been successfully applied in the I2I. By constraining features from the same location to be closer than those from different ones, it implicitly ensures the result to take content from the source. However, previous work uses the features from random locations to impose the constraint, which may not be appropriate since some locations contain less information of source domain. Moreover, the feature itself does not reflect the relation with others. This paper deals with these problems by intentionally selecting significant anchor points for contrastive learning. We design a query-selected attention (QS-Attn) module, which compares feature distances in the source domain, giving an attention matrix with a probability distribution in each row. Then we select queries according to their measurement of significance, computed from the distribution. The selected ones are regarded as anchors for contrastive loss. At the same time, the reduced attention matrix is employed to route features in both domains, so that source relations maintain in the synthesis. We validate our proposed method in three different I2I datasets, showing that it increases the image quality without adding learnable parameters. Codes are available at \href{https://github.com/sapphire497/query-selected-attention}{https://github.com/sapphire497/query-selected-attention}.
\noindent\footnotetext{Corresponding author, email: sunli@ee.ecnu.edu.cn. This work is supported by the Science and Technology Commission of Shanghai Municipality (No.19511120800) and Natural Science Foundation of China (No.61302125). }
\end{abstract}

\begin{figure}[t]
\centering
\includegraphics[width=.9\columnwidth]{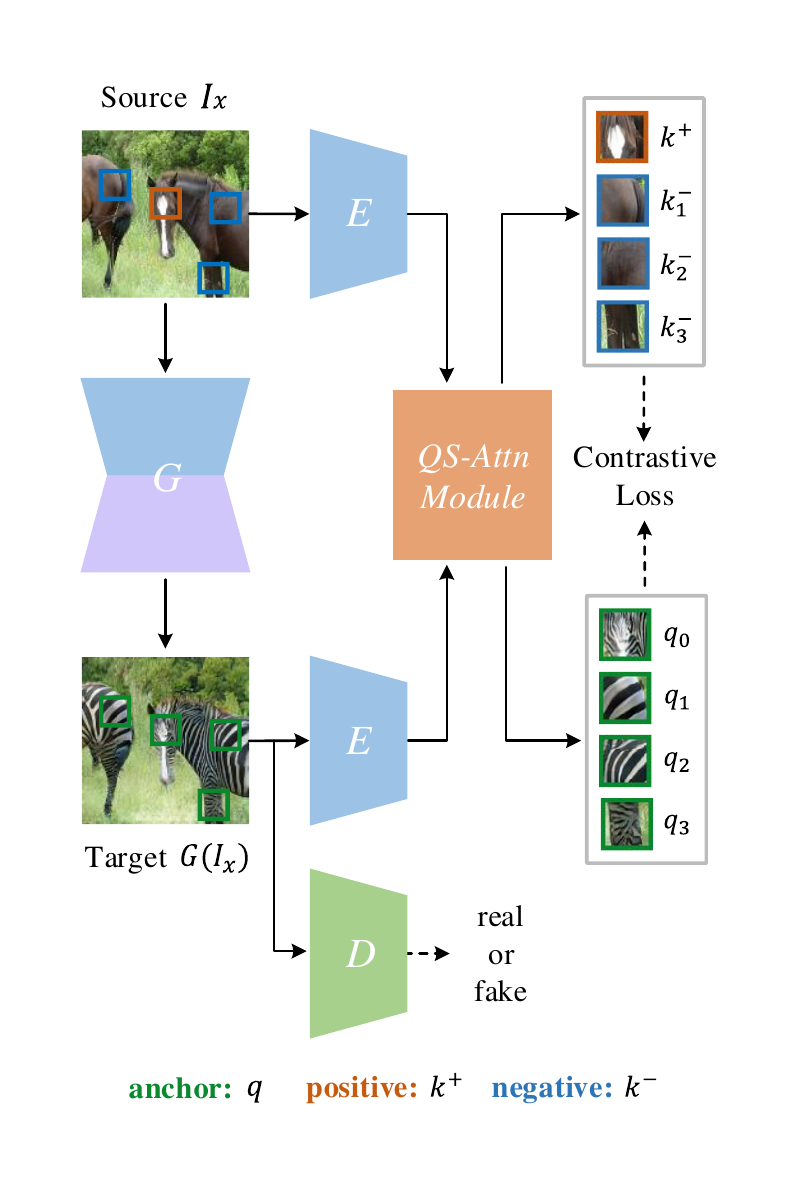}
\caption{The overall structure of our model. The source domain image $I_x$ is translated by the generator $G$ into a target domain image $G(I_x)$. The encoder $E$ extracts features from these two images, then the $QS$-$Attn$ module selects significant features to establish the contrastive loss. We also use a discriminator $D$ to construct the adversarial loss.}
\label{fig:fig1}
\vspace{-0.2cm}
\end{figure}

\begin{figure}[t]
\centering
\includegraphics[width=1\columnwidth]{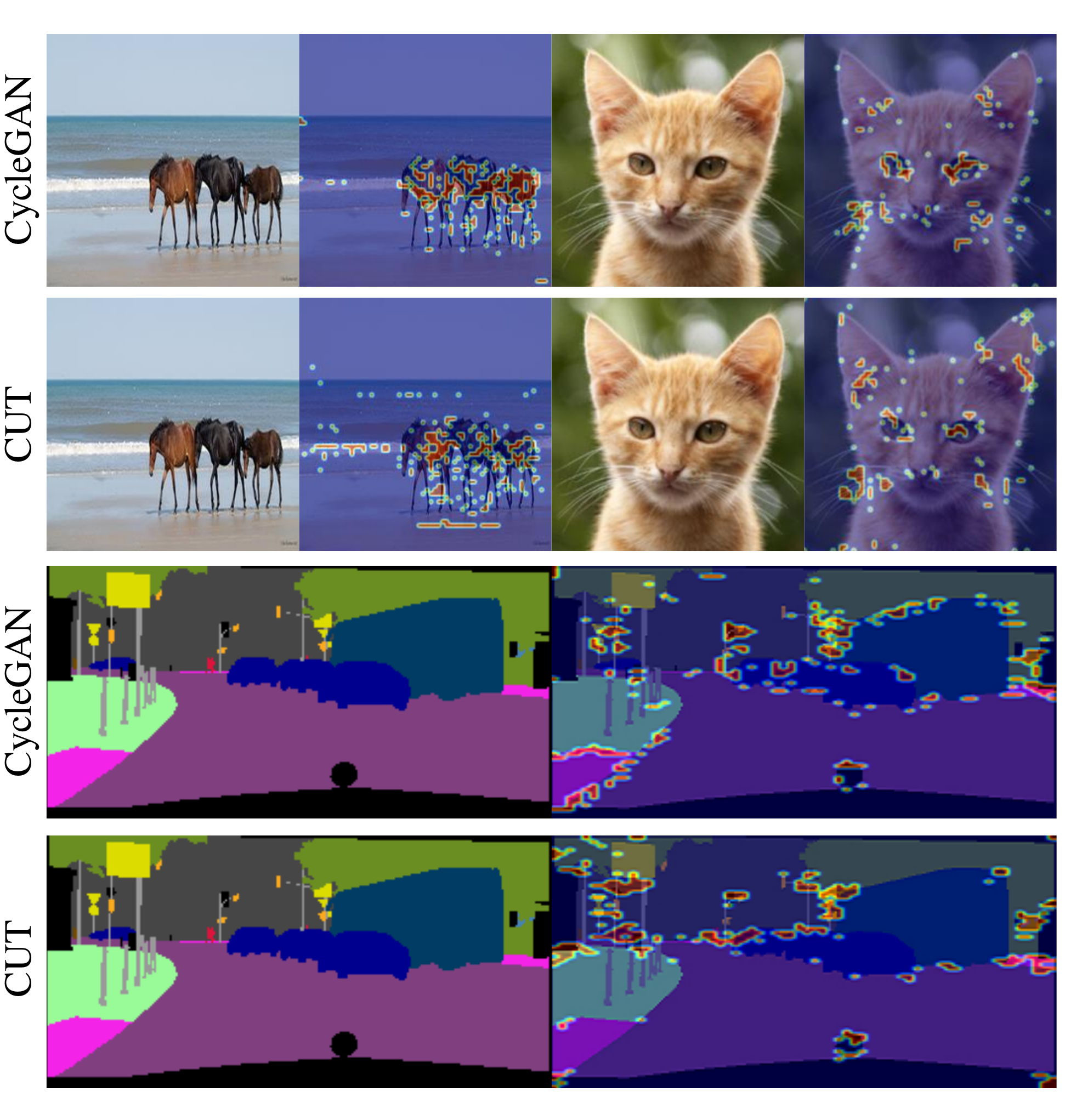}
\caption{Visualization of feature significance metric from pretrained CycleGAN and CUT on three datasets. We show the entropy of attention matrix for each location, the warmer color indicates that the entropy is smaller. For each dataset, on the left column are the input images, and on the right column are the entropy visualizations of two models.}
\label{fig:entropy}
\vspace{-0.1cm}
\end{figure}

\section{Introduction}
In image-to-image (I2I) translation, an input from the source domain $X$ is mapped into the target domain $Y$ while keeping its original content from unnecessary modifications. The translation is usually achieved by a generator $G$ in the structure of auto-encoder with its output constraining by a discriminator $D$, so that it fulfills the requirement of the domain $Y$. In many I2I tasks, paired data are impossible to obtain, hence $G$ can not be directly guided by the real image in $Y$. Ensuring that the output takes the input content is important for increasing its quality. Typical methods proposed in \cite{zhu2017unpaired,yi2017dualgan,kim2017learning} translate the result back into domain $X$ by another generator $G'$, and set up the cycle consistency penalty between the input and the final output. Although improving the quality, they introduce two generators and discriminators, which greatly increase the training costs. 

Recently, CUT \cite{park2020contrastive} incorporates the contrastive learning between the cross domain features from $G$'s output and input. The key idea is to constrain the features from encoder $E$, requiring those from the same location to be close, compared with those from different locations. Removing the \textit{QS-Attn module}, \cref{fig:fig1} illustrates the overall structure of CUT. An anchor point at a random position is selected from the features of the translated image, then one corresponding positive and many negatives are also sampled from the features of input. The contrastive loss is computed for the anchor so that the model maximizes the mutual information between the corresponding features. Note that CUT has only a single direction. Therefore, only one $G$ is needed, and the training cost is reduced. The image quality is greatly improved, showing that contrastive loss across domains is useful in I2I.

However, there are still two issues ignored by CUT, which can be potentially improved. First, it does not select the anchors with purpose in the contrastive learning. Since each of them represents a small patch in the original image resolution and many of them may not reflect any domain characteristics relevant for I2I. We argue that only those containing significant domain information need to be edited, and the contrastive loss imposed on them are more meaningful to guarantee the consistency across domains. Second, each anchor feature has only limited receptive field, and it does not consider its relation with other locations. This relation provides valuable cues to keep the source content stable and make the translation relevant. 

We consider the above two issues in a simple way, inserting the \textit{QS-Attn module} into the model as \cref{fig:fig1} shows, without bringing in extra model parameters. To evaluate the feature significance at different locations, we directly utilize features from $E$ as both queries and keys to calculate the attention matrix in the source domain, then the distribution entropy is computed as a metric. Intuitive illustrations are provided in \cref{fig:entropy}, in which such an entropy metric is visualized in the form of heat map. Particularly, given input images that need to be translated, we apply the encoder of pretrained CycleGAN \cite{zhu2017unpaired} and CUT \cite{park2020contrastive} models to obtain the features and calculate the attention matrix, and then the entropy is computed for each row of it. We sort the entropy in ascending order and show the smallest $N$ points on the image. For the \textit{Horse} and \textit{Cat} images, the entropy values on the body of horse and the face of cat are smaller. For the \textit{Label} image, the points mainly locate at the edges of categories. Consequently, the entropy can be a metric to measure how important the feature is in reflecting domain characteristics, hence we can impose the contrastive loss on it, ensuring the accurate translation on the domain-relevant features.

This paper intends to quantitatively measure the significance of each anchor feature and select the relevant ones for the contrastive loss according to the metric. Based on the previous analysis, we calculate the entropy of each row in the attention matrix and keep those with smaller entropy values. The remaining rows form the query-selected attention (QS-Attn) matrix, which consists of fewer queries, and they are further employed to route the value feature. Here the same matrix is multiplied with the values from both source and target domains, which implicitly  keeps the feature relation in the source domain, avoiding excessive modifications on the result.

The contributions of this paper lie in following aspects:
\begin{itemize}
\item We propose a QS-Attn mechanism in I2I task. Our scheme is to choose the relevant anchor points, and use them as queries to attend and absorb features at other locations, forming better features suitable for contrastive learning. The QS-Attn keeps the simple design in CUT, and does not add any model parameters.
\item We investigate different ways to quantify the significance of the queries, to perform the attention, and to route value features in QS-Attn module, and find the entropy-based measurement and global attention for the cross domain value routing is the robust one.
\item We do intensive experiments on the commonly used datasets, and show the proposed method achieves SOTA in most two domains I2I tasks.
\end{itemize}

\section{Related Works}
\textbf{Image-to-image translation.} GAN \cite{goodfellow2014generative,mirza2014conditional,brock2018large,karras2019style} has the strong ability to describe the high dimensional distribution, therefore, has been widely used in the tasks of image synthesis like super-resolution \cite{ledig2017photo}, de-noising \cite{chen2018image} and I2I. The I2I is first presented in Pix2Pix \cite{isola2017image}, and extended to high-resolution in Pix2PixHD \cite{wang2018high}, which can be regarded as a type of conditional GAN. The generator $G$ consists of a pair of connected encoder-decoder, which translates an image from the source to the target domain. It is trained by paired data together with the adversarial loss from a target domain discriminator. However, unpaired I2I is more desirable, since the matched data across domains are impossible to be collected in most settings. CycleGAN \cite{zhu2017unpaired} and DiscoGAN \cite{kim2017learning} achieve the I2I based on the unpaired data. They simultaneously train two different $G$, being responsible for the two directions of image translation. The cycle consistency is set up by successively using the two $G$ for the opposite translation, and requiring the output to reconstruct the input source, which ensures $G$ to employ the given content during translation, and maximizes the mutual information between the output and input source. The idea can be applied to multiple domains defined by several attributes \cite{choi2018stargan}, or utilized in the feature space \cite{hoffman2018cycada, choi2020starganv2} in a flexible way. Meanwhile, many works \cite{liu2017unsupervised, huang2018multimodal, lee2018diverse} try to give the diverse translations by mixing the content and style from different images, or supporting the random sampling in the latent space. Together with the cycle consistency, translated images with the same content can appear different styles.

However, the cycle consistency is usually blamed for its strong constraint directly on the pixel \cite{nizan2020breaking,zhao2020unpaired}, which is not only unnecessary, but sometimes degrades the image quality. Except computing it in the feature level, another simple way is to perform the single direction translation. In this setting, the key issue becomes to keep the input content, and the extra loss term needs to be added. DistanceGAN \cite{benaim2017one} requires the generator to preserve the pixel level distance across two domains. GcGAN \cite{fu2019geometry} links the source and target through a predefined geometry function. On the other hand, the feature level perceptual loss \cite{johnson2016perceptual,mechrez2018contextual} specified by a pretrained VGG \cite{simonyan2014very} is widely adopted \cite{gatys2016image,chen2017photographic, mechrez2018contextual}, which maintains the high-level semantic of the result. Nevertheless, features from a fixed layer of a pretrained model may not reflect the content which needs to be kept. AttnGAN \cite{chen2018attention} and GANimation \cite{pumarola2018ganimation} learn a foreground mask to guide the generator so that it realizes the translation in the relevant area. But they need extra parameters to estimate the foreground which definitely increase the model complexity. CUT \cite{park2020contrastive} is the first attempt to incorporate the self-supervised contrastive loss into I2I, which significantly increases the translation quality. F-LSeSim \cite{zheng2021spatiallycorrelative} extends CUT by computing the self similarity within a local region, and imposes contrastive loss on it. However, it relies on features from VGG to measure the similarities, which reduces the training efficiency. We emphasize that both CUT and F-LSeSim do not intentionally select anchor features for contrastive loss, and their features still lacks large receptive field for representing image in source domain.

\begin{figure*}[ht]
\centering
\includegraphics[width=0.95\textwidth]{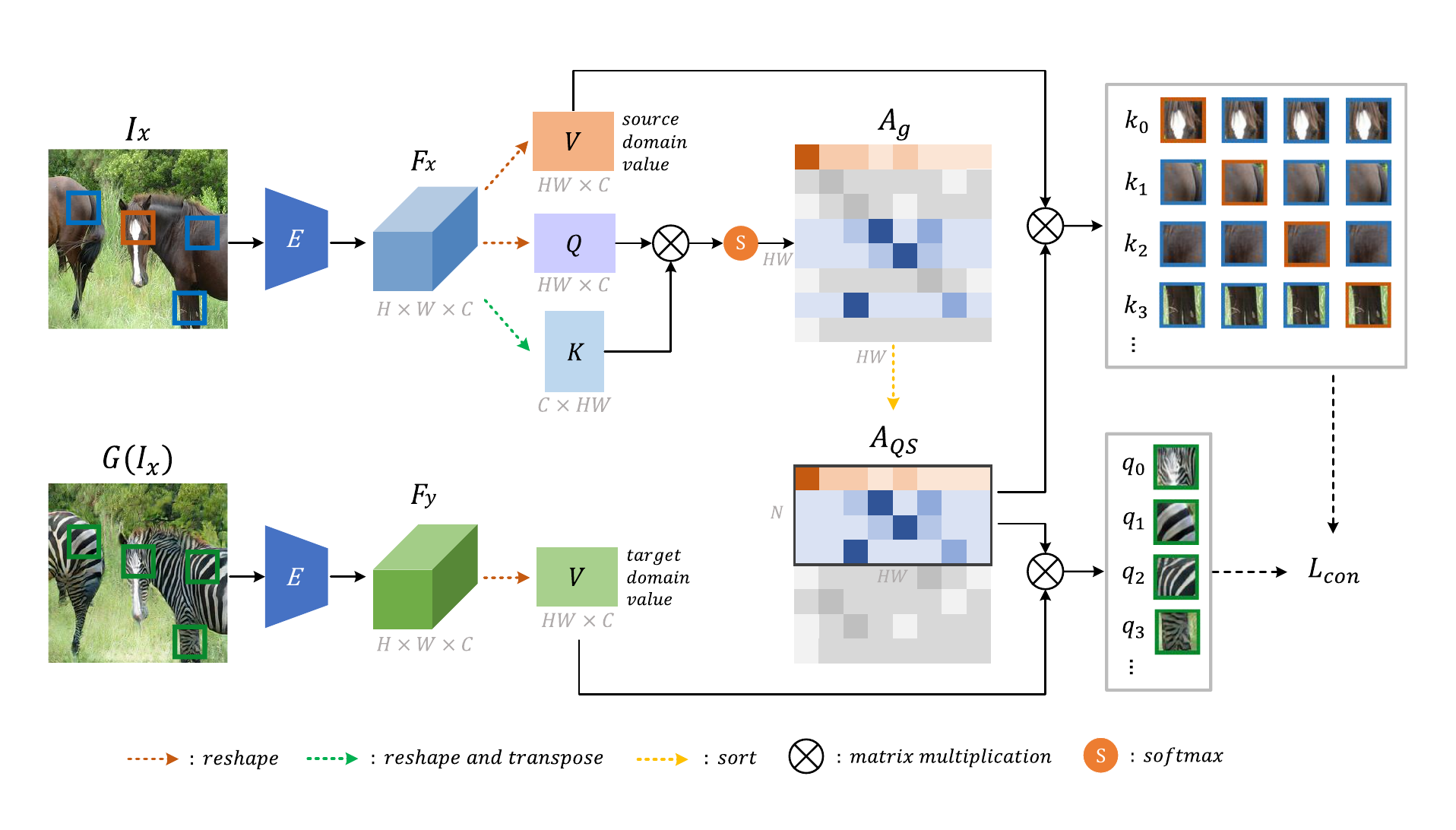}
\vspace{-0.4cm}
\caption{The details of QS-Attn. The encoder $E$ extracts features $F_x$ and $F_y$ from $I_x$ and $G(I_x)$, then $F_x$ is reshaped and computed to derive the attention matrix $A_g$. Each row in $A_g$ is sorted by its metric of the significance, and the selected $N$ rows forming the $A_{QS}$. We further apply $A_{QS}$ to route both source and target domain value features, and obtain positive, negative and anchor features to construct the contrastive loss $L_{con}$. Positive and negatives are from the real image $I_x$, while anchors are from the translated image $G(I_x)$. The patches in orange, blue and green indicate the positive, negative and anchor, respectively.}
\label{fig:fig2}
\end{figure*}

\textbf{Self-supervised contrastive learning.} Despite the great success of supervised learning, deep neural network is accused of its requirement on the large amount of the labeled training data. Recent studies of the self-supervised learning show its strong ability to represent an image without labels, particularly with the help of the contrastive loss \cite{oord2018representation, wu2018unsupervised}. Its idea is to perform the instance level discrimination and learn the feature embedding, by pulling the features from the same image together and pushing those from different ones away. Recently, it has been investigated as a pretraining technique \cite{chen2020simple,he2020momentum,chen2020improved,grill2020bootstrap,caron2020unsupervised,chen2020exploring}, providing the initial model or the latent embedding for the down-stream task. The self-supervised learning has also been applied in image generation \cite{chen2019self,patel2021lt}. SS-GAN \cite{chen2019self} incorporates the rotation degree prediction as an auxiliary task for the discriminator, preventing the overfitting due to the limited data for real/fake binary classification. LT-GAN \cite{patel2021lt} trains an auxiliary classifier on top of the embedding in the discriminator, to classify whether the two pairs of fake images have the same perturbations on the sampling noise vectors. Besides CUT \cite{park2020contrastive}, the work \cite{kang2020online} also adopts the self-contrastive learning for I2I. It employs non-local attention matrix to warp target image to the source pose, and requires feature from warped image to be close to the source through the constrastive loss. None of them chooses anchors, or utilizes the relation in the source domain like our method. 

\section{Methods}
\subsection{Preliminaries on CUT}
In the task of I2I, given an image $I_x \in \mathbb{R}^{H\times W\times 3}$ from source domain $X$, the model aims to translate it into $G(I_x)$ in the target domain $Y$, having no obvious distinctions with the real image $I_y \in \mathbb{R}^{H\times W\times 3}$ in that domain. Generally, there are two auto-encoders, one $G_{X\rightarrow Y}$ for $X$ to $Y$ and the other $G_{Y\rightarrow X}$ for the reverse. CUT focuses on the single-direction translation from $X$ to $Y$, so it only needs one generator $G$ and one discriminator $D$, therefore, the subscript is omitted. The objective of adversarial loss $L_{adv}$ can be computed as follows.
\begin{equation}
\label{eq:loss_adv}
\begin{aligned}
	L_{adv} =  \mathbb{E}_{I_y\in Y} \log \text{D}(I_y) + \mathbb{E}_{I_x\in X} \log (1 - \text{D}(\text{G}(I_x))
\end{aligned}
\end{equation}

Besides the $L_{adv}$ in \cref{eq:loss_adv}, CUT takes advantage of the first half of $G$ as an encoder $E$ to provide the extra constraint for the output from $G$. Basically, $E$ compares the feature similarities across different domains. This efficient scheme is slightly different from previous work \cite{zhu2017toward}, in which another encoder $E$ is employed, and is also adopted in \cite{liu2021divco}. It extracts features from both $I_x$ and $G(I_x)$, and establishes the self-supervised contrastive loss in \cref{eq:loss_con}, 
\begin{equation}
\label{eq:loss_con}
\begin{aligned}
	L_{con}=-\log \left[ \frac{\exp({q}\cdot{k^+}/\tau)}{\exp({q}\cdot{k^+}/\tau) + \sum_{i=1}^{N-1}\exp({q}\cdot{k^-}/\tau)}\right]
\end{aligned}
\end{equation}
where $q$ is the anchor feature from $G(I_x)$, $k^+$ is a single positive and $k^-$ are $(N-1)$ negatives. Here $\tau$ indicates a temperature hyper-parameter. Note that the anchor $q$ always locates in the fake image, and its positive $k^{+}$ is on the same location in the real image $I_x$. Besides, $(N-1)$ negatives $k^{-}$ are randomly selected in $I_x$. The gradient of $L_{con}$ only applies on the anchor $q$ to train the parameters in $G$, while it is detached on $k^+$ and $k^-$, so that $G$ is guided for the single direction of domain translation.

The full objective is expressed as follows.
\begin{equation}
\label{eq:loss_full}
\begin{aligned}
	L_{G} = L_{adv} + L_{con}^X + L_{con}^Y
\end{aligned}
\end{equation}
where $L_{con}^X$ is the contrastive loss defined in \cref{eq:loss_con}, and $L_{con}^Y$ is the identity loss, in which the positive $k^+$ and negatives $k^-$ are from a real target domain image $I_y$, and the anchor $q$ is from $G(I_y)$. This identity loss guarantees the features from $G(I_y)$ are similar with features from $I_y$, preventing $G$ from making changes on the target domain images.

\subsection{QS-Attn for Contrastive Learning}
As is shown in \cref{fig:fig1}, we keep the simple setting like CUT. $E$ is applied to extract features from $I_x$ and $G(I_x)$. These features are supposed to establish the contrastive loss $L_{con}$ defined in \cref{eq:loss_con}. The key module QS-Attn, setting up $L_{con}$ across two domains, is illustrated in \cref{fig:fig2}. Instead of the simple random strategy, we employ the idea of attention, which first compares a given query with keys, and then selects the query based on the comparison result. However, we do not use any separated projection head for query, key and value like the common attention, therefore, without adding extra model parameters in both $G$ and $D$. Details for QS-Attn are given in the following two sub-sections. 
\subsubsection{Attention for query selection.} 
CUT randomly selects the anchor $q$, positive $k^+$ and negatives $k^-$ to compute the contrastive loss in \cref{eq:loss_con}, which is potentially inefficient, because their corresponding patches may not come from the domain-relevant region, \emph{e.g.} the horse body in the \textit{Horse $\rightarrow$ Zebra} task. Note that some features do not reflect the domain characteristics, they tend to be kept during the translation. Therefore, the $L_{con}$ imposed on them is not vital for $G$. Our intention is to choose the anchor $q$, and compute $L_{con}$ on the significant ones which contains more domain-specific information.

\indent\textbf{Global attention. }
Based on the above observation, we aim to define a quantitative value for each potential location, which reflects the significance of the feature. The quadratic attention matrix is adopted, since it exhaustively compares each feature with all other locations, it accurately reflects the similarities with others, as is shown in \cref{fig:fig2}. Particularly, given a feature $F_x\in \mathbb{R}^{H\times W\times C}$ in the source domain, we first reshape it into a 2D matrix $Q\in\mathbb{R}^{HW\times C}$, and then multiply it by its transposed $K\in \mathbb{R}^{C\times HW}$. Then we give each row of the matrix to the softmax function, leading to a global attention matrix $A_g\in\mathbb{R}^{HW\times HW}$. Consequently, significant features can be measured according to the entropy $H_g$ of each row in $A_g$, which is computed as in \cref{eq:entropy}.
\begin{equation}
\label{eq:entropy}
H_{g}(i)=-\sum_{j=1}^{HW} A_g(i,j)\log A_g(i,j)
\end{equation}

Here $i$ and $j$ are the indexes of the query and key, corresponding to the row and column in $A_g$. When $H_{g}(i)$ approaches to 0, it means that in the $i$-th row, only a very few key locations are similar with the $i$-th query. Hence we assume that it is distinct enough and is important to be constrained by $L_{con}$. To select all the significant queries, the rows of $A_g$ are sorted by the entropy $H_g$ in the ascending order, and the smallest $N$ rows are selected as the QS-Attn matrix $A_{QS}\in \mathbb{R}^{N\times HW}$. Note that $A_{QS}$ is fully determined by the features in $I_x$, and has no relation with $G(I_x)$. 

\indent\textbf{Local attention. }
Though non-local attention can obtain the global context, it smooths out the detailed context surrounding the queries. Local attention measures the similarity between a query and its neighboring keys within a constant window of $w\times w$ and stride of $1$, which can capture the spatial interactions in local regions, and reduce the computation cost. Given a reshaped query matrix $Q_{l}\in\mathbb{R}^{HW\times C}$, we multiply it by local key matrix $K_{l}\in\mathbb{R}^{HW\times w^2\times C}$ and send it to the softmax function, leading to a local attention matrix $A_l\in\mathbb{R}^{HW\times w^2}$. The local entropy $H_{l}$ is computed in each row as in \cref{eq:localentropy}.
\begin{equation}
\label{eq:localentropy}
H_{l}(i)=-\sum_{j=1}^{w^2} A_l(i,j)\log A_l(i,j)
\end{equation}
Here $i$ and $j$ are the indexes of the query and key. We select the smallest $N$ rows in $A_l$ by sorting $H_{l}$ in the ascending order to form $A_{QS}$. For the value routing, We also locate the $N$ indexes in local value matrix $V_{l}\in\mathbb{R}^{HW\times w^2\times C}$ and get the selected value matrix $V_{ls}\in\mathbb{R}^{N\times w^2\times C}$. 

\subsubsection{Cross domain value routing for contrastive learning.} 
The reduced $A_{QS}$ is used as the attention matrix to route the value features from both source and target domains. Here we emphasize that $A_{QS}$ captures the global or local relation by comparing the query with keys, and it provides useful high-order descriptions about the shape and texture of $I_x$. Using it to route features helps to enlarge the receptive field of the selected queries, so that better features, which consider the context of $I_x$, can be formulated. Moreover, the relation defined by $A_{QS}$ is also required to be kept during the image translation. So $A_{QS}$ is imposed on the features from both $I_x$ and $G(I_x)$, routing the corresponding value to form the anchor, positive and negatives. One positive and $(N-1)$ negative features are located in the real image $I_x$. $N$ anchors are from the fake image $G(I_x)$. We establish the self-supervised contrastive loss as \cref{eq:loss_con}, using these features to constrain the translation. 

\begin{figure*}[ht]
\centering
\includegraphics[width=1\textwidth]{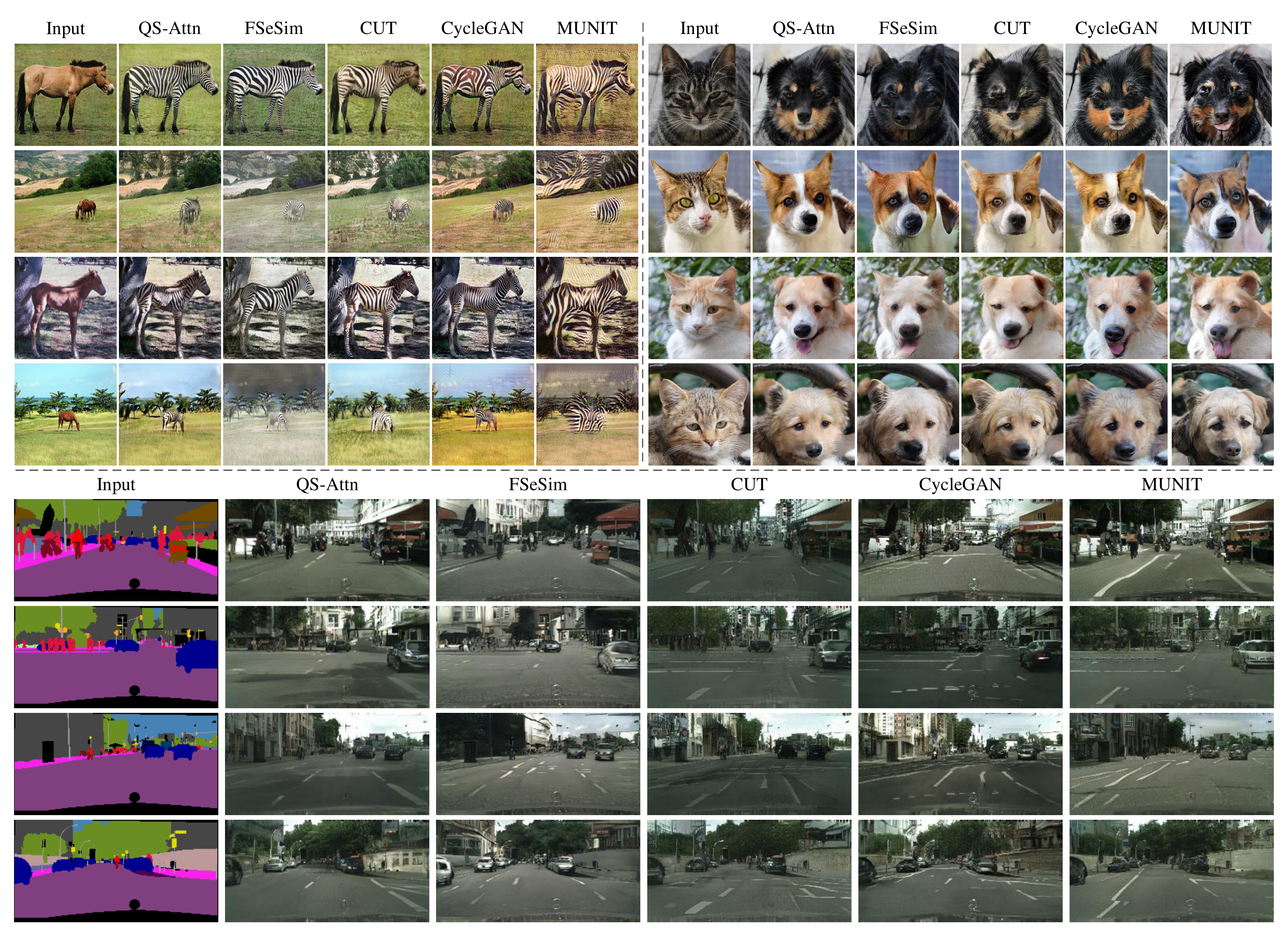}
\vspace{-0.4cm}
\caption{Visual results comparison with other methods. We compare our model with FSeSim, CUT, CycleGAN and MUNIT on three benchmark datasets. In the results of QS-Attn, the translated images of \textit{Horse $\rightarrow$ Zebra} and \textit{Cat $\rightarrow$ Dog} are from QS-Attn(Global), and the results of \textit{Cityscapes} are generated using QS-Attn(Global+Local). 
	More results can be found in the appendix.}
\label{fig:baseline}
\vspace{-0.1cm}
\end{figure*}

\begin{table*}[t]
\centering
\begin{tabular}{lcccccccc}
\toprule
\multirow{2}{*}{\textbf{Method}}  & \multicolumn{4}{c}{\textbf{Cityscapes}} & \multicolumn{2}{c}{\textbf{Cat$\rightarrow$Dog}} & \multicolumn{2}{c}{\textbf{ Horse$\rightarrow$Zebra}} \\ 
\cmidrule(l){2-5} \cmidrule(l){6-7} \cmidrule(l){8-9}
& \textbf{mAP}$\uparrow$ & \textbf{pixAcc}$\uparrow$ & \textbf{classAcc}$\uparrow$ & \textbf{FID}$\downarrow$ & \textbf{SWD}$\downarrow$ & \textbf{FID}$\downarrow$   & \textbf{SWD}$\downarrow$ & \textbf{FID}$\downarrow$\\
\midrule
{CycleGAN} & 20.4 & 55.9 & 25.4 & 76.3 & 19.5 & 85.9 & 39.1 & 77.2 \\ 
{MUNIT} & 16.9 & 56.5 & 22.5 & 91.4 & 24.4 & 104.4 & 50.7 & 133.8\\
{CUT} & 24.7 & 68.8 & 30.7 & 56.4 & 12.9 & 76.2 & 31.5 & 45.5\\
{FSeSim} & 22.1 & 69.4 & 27.8 & 54.3 & 13.8 & 87.8 & 37.2 & 43.4\\ \cdashline{1-9} 
{QS-Attn(Global)} & 25.5  & 79.9 & 31.2 & 53.5 & \textbf{12.8} & \textbf{72.8} & \textbf{30.3} & 41.1 \\ 
{QS-Attn(Local)} & 26.2 & 80.5 & 31.9 & \textbf{48.8} & 13.3 & 79.3 & 31.2 & \textbf{38.6}\\ 
{QS-Attn(Local+Global)}  & \textbf{27.9} & \textbf{81.4} & \textbf{32.6} & 50.2 & 13.2 & 80.0 & 31.9 & 42.3\\
\bottomrule
\end{tabular}
\vspace{-0.1cm}
\caption{Quantitative comparison with other methods. The last three rows are our models with different settings, details are illustrated in \cref{subsec:results}. The best performance is indicated in \textbf{bold}.
}
\vspace{-0.3cm}
\label{tab:tab1}
\end{table*}

\begin{figure*}[ht]
\centering
\includegraphics[width=.95\textwidth]{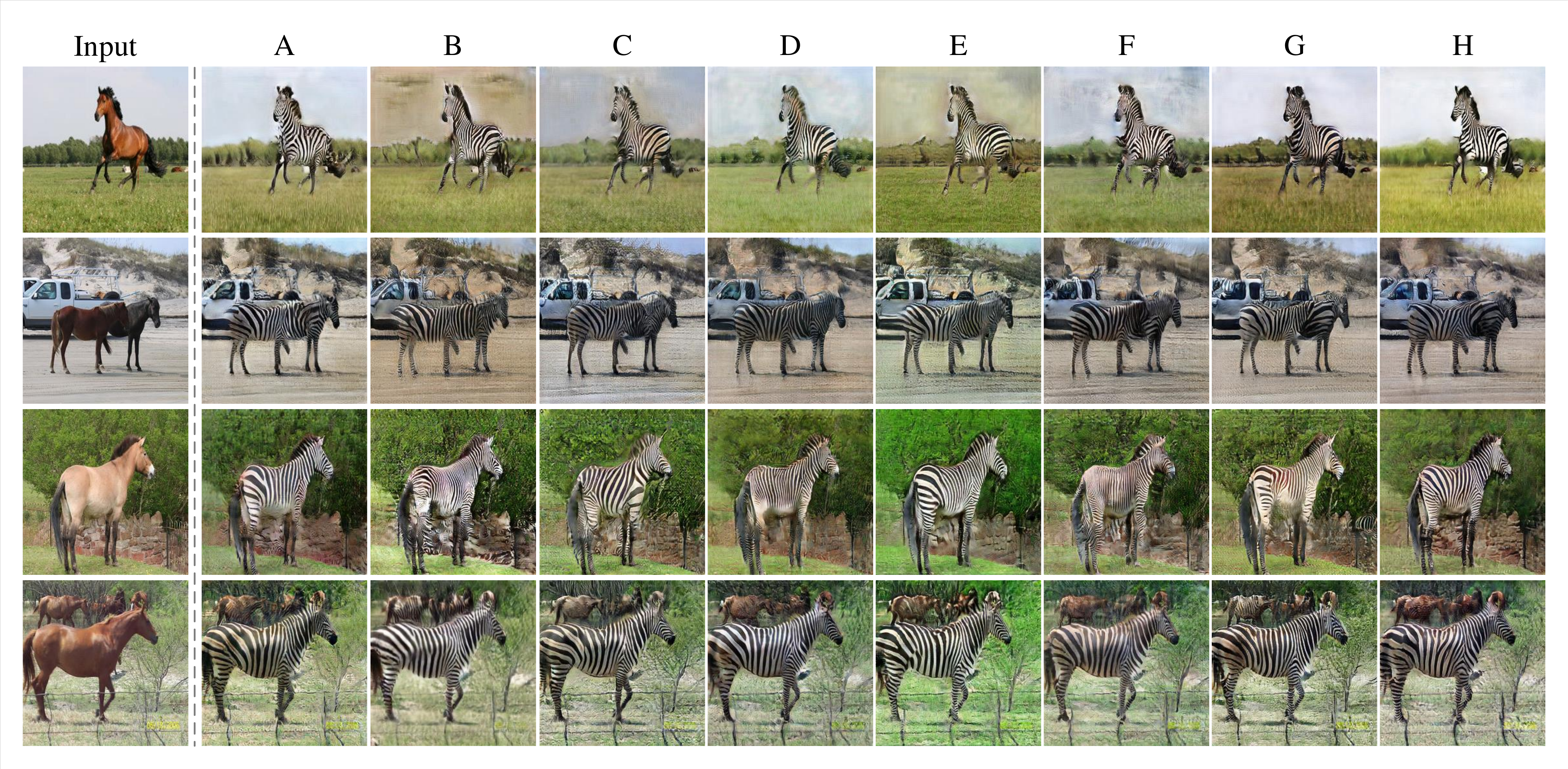}
\vspace{-0.4cm}
\caption{Qualitative ablation results. The leftmost column are input images, the remaining columns are translated images using model A-H. Details are illustrated in \cref{tab:ablation} and \cref{subsec:ablation}.}
\label{fig:ablation}
\vspace{-0.3cm}
\end{figure*}

\section{Experiments}
\subsection{Implementation Details}
\indent\textbf{Datasets.}
Our model is trained and evaluated on \textit{Cityscapes}, \textit{Horse $\rightarrow$ Zebra}, and \textit{Cat $\rightarrow$ Dog} datasets. \textit{Cityscapes} contains street scenes from German cities, with 2,975 training images. \textit{Cat $\rightarrow$ Dog} are from AFHQ Dataset \cite{choi2020starganv2}, which consists of 5,153 and 4,739 training images for cat and dog, respectively. The images for \textit{Horse $\rightarrow$ Zebra} are provided in \cite{zhu2017unpaired}, which contains 1,067 and 1,334 training images for horse and zebra, respectively. For all experiments, the resolution of input and generated images is $256\times256$. The initial resolution of images in \textit{Cityscapes} and \textit{Cat $\rightarrow$ Dog} dataset are $2048\times1024$ and $512\times512$, and we resize them to $256\times256$ in our experiments.

\indent\textbf{Training details.}
We build our model with a ResNet-based generator and a PatchGAN discriminator \cite{son2017retinal}, and compare it with CUT in the same setting on three aforementioned datasets. The number of rows in attention matrix is set to 256, and the dimension of anchor, query, key features for computing contrastive loss is 256. We adopt the multi-layer feature extraction in CUT, which takes the features from five layers. Considering the high cost in computing the global attention matrix, we propose to apply the QS-Attn on the last two layers’ feature in the encoder, but we still make an extra comparison with the model in which all layers are applied it. Details can be found in the supplementary materials.

\indent\textbf{Evaluation metrics.} 
We use Fr\'echet Inception Distance (FID) \cite{NIPS2017_8a1d6947} and Sliced Wasserstein Distance (SWD) \cite{rabin2011wasserstein} to evaluate the quality of translated images. FID and SWD both measure the distance between two distributions of real and generated images, and lower indicate the generated image is similar to the real one. For \textit{Cityscapes} dataset, following \cite{park2020contrastive}, we apply semantic segmentation to the generated images using DRN \cite{yu2017dilated}, and compute mean average precision (mAP), pixel-wise accuracy (pixAcc), and average class accuracy (classAcc), showing the semantic interpretability of the generated images. We calculate the metrics in the whole test set, in which \textit{Cityscapes} contains 500 label images, \textit{Cat $\rightarrow$ Dog} includes 500 cat images, and \textit{Horse $\rightarrow$ Zebra} comprises 120 horse images.

\subsection{Results}\label{subsec:results}
\indent\textbf{Quantitative and qualitative results.}
\cref{tab:tab1} compares our model with FSeSim \cite{zheng2021spatiallycorrelative}, CUT, CycleGAN \cite{zhu2017unpaired} and MUNIT \cite{huang2018multimodal}. There are three settings in our model, \textit{Global}, \textit{Local} and \textit{Local+Global}. \textit{Global} means sorting the global attention matrix $A_g$ by entropy to form the QS-Attn matrix $A_{QS}$, and \textit{Local} means sorting the local attention matrix $A_l$ to form $A_{QS}$. Furthermore, in order to utilize both local and global context, \textit{Local+Global} applies $A_l$ for query selection and $A_g$ in value routing. The rows of $A_g$ are sorted by the local entropy $H^{l}$ in the ascending order, and the smallest $N$ rows are selected as $A_{QS}$. The ablation study of the three models is illustrated in \cref{subsec:ablation} and the code is provided in the supplementary materials. For the metric of FID, the translated results of our model are more realistic than other methods on three datasets. Our model also performs better on mAP, pixAcc and classAcc on \textit{Cityscapes} dataset. Moreover, our method \textbf{does not} add extra model parameters in both $G$ and $D$, and uses the same architecture of $G$ as CycleGAN. 

Visual results are shown in Figure \ref{fig:baseline}. Compared to other methods, our QS-Attn model has the ability to translate the domain-relevant features accurately. Besides, QS-Attn achieves the background consistency in the tasks of \textit{Horse $\rightarrow$ Zebra} and \textit{Cat $\rightarrow$ Dog}.

\begin{table}[t]
\centering
\begin{tabular}{@{}ccccc@{}}
\toprule
& \textbf{QS-Attn} & \textbf{FSeSim} & \textbf{CUT} & \textbf{CycleGAN} \\ \midrule
\textbf{Q(\%)} & \textbf{45.0}    & 28.3            & 13.3         & 13.3              \\
\textbf{T(\%)} & \textbf{40.0}    & 30.0            & 11.7         & 18.3              \\
\textbf{C(\%)} & \textbf{58.3}    & 11.7            & 8.3          & 21.7              \\ \bottomrule
\end{tabular}
\vspace{-0.1cm}
\caption{User study statistics. The methods are compared in three aspects: image quality (\textit{Q}), target domain conformity (\textit{T}) and domain-irrelevant consistency (\textit{C}).}
\label{tab:user}
\vspace{-0.3cm}
\end{table}

\begin{table*}[ht]
\centering
\begin{tabular}{lccccccc}
\toprule
\multicolumn{1}{c}{\multirow{3}{*}{\textbf{Method}}} & \multicolumn{5}{c}{\textbf{Configuration}} & \multirow{3}{*}{\textbf{FID $\downarrow$}} & \multirow{3}{*}{\textbf{ SWD $\downarrow$}}\\  \cmidrule(r){2-6}
\multicolumn{1}{c}{} & \multicolumn{2}{c}{\textbf{Attention}} & \multicolumn{2}{c}{\textbf{Selection}}   & \multirow{2}{*}{\textbf{Cross Domain}}& \\ 
\cmidrule(r){2-3} \cmidrule(r){4-5}
\multicolumn{1}{c}{} & \textbf{global}   & \textbf{local}   & \textbf{global}  & \textbf{local} &   \\ \midrule
\textbf{A}  & $\checkmark$  &  &$\checkmark$ &  & $\checkmark$  & 41.1 & \textbf{30.3}\\
\textbf{B}  & $\checkmark$  &  & $\times$    &  & $\checkmark$   & 52.1 & 34.4    \\
\textbf{C}  & $\times$      &  & $\checkmark$&  & $\times$    & 61.1 & 37.7          \\ \cdashline{1-8}
\textbf{D}  & $\checkmark$  &  & $\checkmark$&   & $\times$  & 43.3 & 32.1          \\
\textbf{E}  & $\checkmark$  &  & $\times$    &   & $\times$  & 53.4 & 34.9          \\ \cdashline{1-8}
\textbf{F}: info & $\checkmark$ & & $\checkmark$ &  & $\checkmark$ & 43.9 & 33.3 \\ \cdashline{1-8}
\textbf{G}  &   &$\checkmark$  &     &$\checkmark$   & $\checkmark$  & \textbf{38.6} & 31.2          \\
\textbf{H}  &$\checkmark$   &  &     &$\checkmark$   & $\checkmark$  & 42.3 &  31.9         \\
\bottomrule
\end{tabular}
\vspace{-0.2cm}
\caption{Quantitative results for ablation study. In configuration, $Attention$ means computing the QS-Attn matrix $A_{QS}$ and adopting it to route the value feature; $Selection$ denotes selecting queries by entropy in attention matrix; $Cross\,Domain$ indicates that $A_{QS}$ in the source domain is applied to route the value features from both source and target domains; $global$ and $local$ refer to using $A_g$ and $A_l$, respectively. Model $\textbf{A}$, $\textbf{G}$ and $\textbf{H}$ are the 3 settings corresponding to the last 3 rows in Table \ref{tab:tab1}.}
\label{tab:ablation}
\vspace{-0.3cm}
\end{table*}

\indent\textbf{User Study.}
To further evaluate the quality of translated images, we conduct a user study under human perception. so a user study reflecting human perception is important for evaluating visual quality. We compare our \textit{Global} model with CUT, FSeSim, and CycleGAN on \textit{Horse $\rightarrow$ Zebra} dataset. 60 participants are asked to compare the methods in three aspects: image quality (\textit{Q}), target domain conformity (\textit{T}) and domain-irrelevant consistency (\textit{C}). \textit{Q} means the reality and perception of images. \textit{T} indicates whether the translated images have the features of the target domain. \textit{C} refers to the domain-irrelevant pixels should remains the same compared to the source images, \emph{e.g.}, the background in the \textit{Horse} image. Statistical results are shown in \cref{tab:user}.

\subsection{Ablation Study}\label{subsec:ablation}
\indent\textbf{Attention and selection. }
In QS-Attn module, we apply attention, query selection and cross domain value routing. To evaluate their effects separately, we conduct the ablation study on \textit{Horse $\rightarrow$ Zebra} dataset. Metrics are listed in \cref{tab:ablation} and qualitative results are shown in \cref{fig:ablation}. $\textbf{A}$ is our complete global model, including aforementioned three operations. In model $\textbf{B}$, queries are selected randomly and the corresponding QS-Attn matrix $A_{QS}$ is applied to route the source and target values. Model $\textbf{C}$ only computes the global attention matrix $A_g$ to select queries, while it does not further use $A_{QS}$ to route values. Model $\textbf{A}$ outperforms model $\textbf{B}$, reflecting the effectiveness of entropy-sorted selection. The metric of $\textbf{C}$ is worse than model $\textbf{A}$ and $\textbf{B}$, indicating that only when the selected $A_{QS}$ routes the values to establish the contrastive loss $L_{con}$, then the encoder learns to extract the significant features from images.

\indent\textbf{Self domain value routing. }
Model $\textbf{D}$ and $\textbf{E}$ in \cref{tab:ablation} route value features in self domain, \emph{i.e.} there are two global attention matrices $A_g^x$ and $A_g^y$, from source domain and target domain, respectively. Then, after the query selection, the source domain QS-Attn matrix $A_{QS}^x$ and the target domain QS-Attn matrix $A_{QS}^y$ route the value features from their own domain. In model $\textbf{D}$, queries are selected by sorting the entropy of $A_g^x$ to form $A_{QS}^x$, and $A_{QS}^y$ is composed of the selected queries in $A_g^y$, which has the same 
row index as $A_g^x$. Different from $\textbf{D}$, model $\textbf{E}$ selects queries randomly. The two models both route the values separately by the corresponding $A_{QS}^x$ and $A_{QS}^y$. Model $\textbf{D}$ is not good as $\textbf{A}$. It demonstrates that compared with self domain routing, cross domain routing can establish closer correlation between source and target domain, for $A_{QS}^x$ captures the global relation of $I_x$ and imposes it on the features from $G(I_x)$.

\indent\textbf{Informer. }
We also investigate other query selection strategies. Recently, Informer \cite{Zhou2020InformerBE} proposes an efficient self-attention mechanism. It introduces a max-mean measurement for each query, which is expressed as:
\begin{equation}
	M(q_{i};K)= \max_{j}\left(\frac{q_{i}k_{j}^T}{\sqrt{C}}\right)-\frac{1}{HW}\sum_{j=1}^{HW}\left(\frac{q_{i}k_{j}^T}{\sqrt{C}}\right)
\end{equation}
where $q_i\in \mathbb{R}^{C}$ is the $i$-th query in the matrix $Q\in \mathbb{R}^{HW\times C}$, $k_j\in \mathbb{R}^{C}$ is the $j$-th key in the matrix $K\in \mathbb{R}^{C\times HW}$. Each query measures the similarity with all keys, and obtains a score $M(q_{i};K)$. Then $Top$-$N$ queries are selected by sorting $M(q_{i};K)$ in descending order, which are relatively distinct. Model \textbf{F} in \cref{tab:ablation} selects $Top$-$N$ queries adopting the above approach, and uses the corresponding $A_{QS}$ to route both source and target domain value features. It achieves good results and metric, showing that the max-mean measurement is also effective for query selection. 

\indent\textbf{Local and global attention. }
Model $\textbf{G}$ and $\textbf{H}$ applies local attention for $A_l$, and computes $H_l$ for query selection. Differently, $\textbf{G}$ employs selected rows in $A_l$ to form $A_{QS}$ in value routing, but $\textbf{H}$ uses global attention matrix $A_g$. Although $\textbf{G}$ achieves the best performance in FID, the visual quality is worse than $\textbf{H}$ and $\textbf{A}$, suggesting that global value routing helps to reconstruct the texture of images.

\section{Conclusion}
This paper proposes a QS-Attn module for cross domain contrastive learning in the task of I2I. Instead of randomly selecting the anchor, positive and negatives to compute the contrastive loss, we measure the significance of source domain features and select them based on a metric so that the constraint becomes more relevant for the domain translation. We first compute an attention matrix using the features from real images in the source domain, and then measure the entropy of every query in it. Those with smaller values are considered to be distinct, therefore being selected. The remaining significant queries are kept, resulting in a row reduced attention matrix, which is further employed to route features in both the source and target domains. The cross domain routing strategy not only enlarges the receptive field of the selected features, but also helps the output to retain the relation in the input image. We show the effectiveness of QS-Attn module on popular domain translation datasets and perform intensive ablation studies. 
\clearpage

\clearpage
\appendix
{\LARGE\noindent\textbf{Appendix}}
\vspace{0.5cm}
\section{Limitations}
We now discuss the limitation mainly existing in the experiments.
Our goal is to select the significant features which are domain-specific 
for contrastive learning, therefore, we choose CUT as a baseline. But we do not 
implement our QS-Attn 
module in other I2I models due to limited training resources, 
such as bi-directional \cite{zhu2017unpaired} and multi-domain I2I tasks \cite{choi2018stargan,choi2020starganv2}, although we think it should also work in them. 
Moreover, the training speed of our model is slower than CUT due to the complex matrix multiplication in global attention, but our model with local attention mitigates this problem, refer to \cref{sec:training} for details.

\section{Additional Ablation Study} \label{sec:abalation}
\begin{figure}[t]
\centering
\includegraphics[width=0.9\columnwidth]{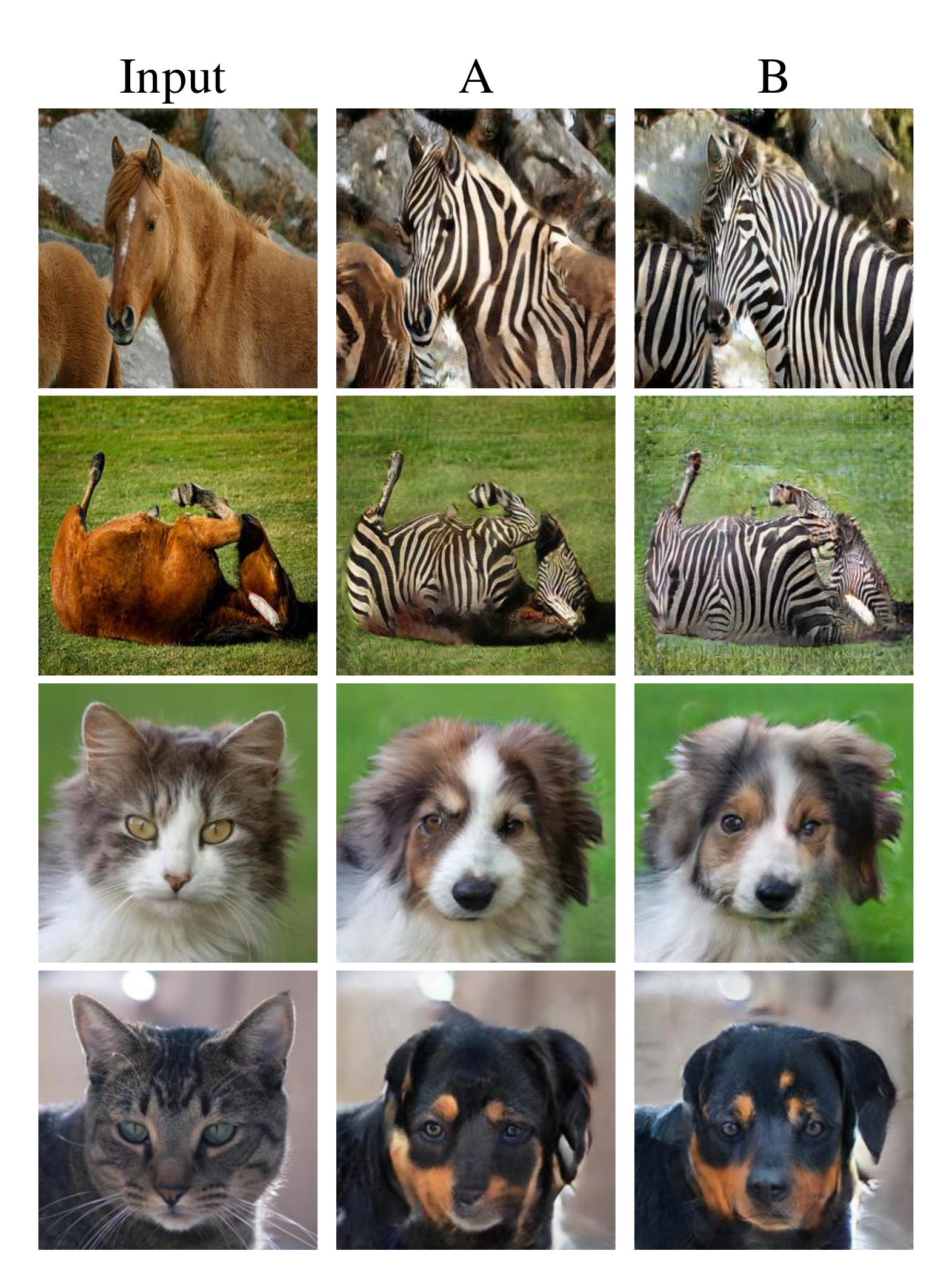}
\caption{Ablation results of different layers for QS-Attn on two datasets.}
\label{fig:appablation}
\end{figure}

In the previous discussion, we placed the emphasis on the query selection and cross domain value routing. Furthermore, in this section, we conduct an additional ablation study of applying our QS-Attn (Global) module in different layers. Quantitative results are illustrated in \cref{tab:appablation}, and visual results are shown in \cref{fig:appablation}. Model \textbf{A} is our initial model, we only employ the QS-Attn on the last two layers, and randomly select points on the first three layers. In model \textbf{B}, we intend to utilize features from all layers. However, the spatial dimension of shallow features is large, hence the computation of global attention matrix $A_g\in\mathbb{R}^{HW\times HW}$ is expensive and time-consuming. Take the high demand of computing into consideration, we employ the average pooling to reduce the spatial size of the feature maps in first three layers. Then, the spatial size of features from all layers is set to $64\times64$, thus $A_g\in\mathbb{R}^{4096\times 4096}$. Model \textbf{B} can compete with model \textbf{A} on the two tasks. Although the average pooling decreases the computing effort, the calculation of $A_g$ and entropy still occupies a great quantity of memory and time. As a result, Our model only exerts the QS-Attn on two layers, which is more lightweight and effective. 

Moreover, We realize that intentionally selecting query may reduce the required number. So we train QS-Attn (Global) with 64, 128 or 512 queries on \textit{Horse $\rightarrow$ Zebra} dataset, and compare its performance with CUT. 
\cref{tab:number} 
demonstrates that the selection of queries is always valid regardless of the number of queries. Moreover, when selecting 128 queries, the performance of QS-Attn is better than which under our original setting. It proves that entropy selection strategy is effective from another perspective.

\section{Query-Selection Algorithm}
There are three query-selection algorithm we proposed: \textit{Global}, \textit{Local}, and \textit{Local+Global}. The pseudo-code in PyTorch style is provided in Algorithm \ref{alg:global},\ref{alg:local} and \ref{alg:localglobal}.

\begin{table}[t]
\label{tab:appablation}
\begin{tabular}{cccccc}
\toprule
\multirow{2}{*}{Method} & \multirow{2}{*}{Layers} & \multicolumn{2}{c}{\textbf{Cat$\rightarrow$Dog}} & \multicolumn{2}{c}{\textbf{ Horse$\rightarrow$Zebra}} \\ \cmidrule(r){3-4} \cmidrule(r){5-6} 
&        & \textbf{SWD$\downarrow$}  & \textbf{FID$\downarrow$}  & \textbf{SWD$\downarrow$}  & \textbf{FID$\downarrow$} \\ \midrule
\textbf{A}              & last 2 & 12.8            & \textbf{72.8}  & \textbf{30.3}    & 41.1 \\
\textbf{B}              & all    & \textbf{12.6}   & 73.1           & 32.5             & \textbf{40.4} \\ 
\bottomrule

\end{tabular}
\caption{Ablation study of different layers for QS-Attn. Model A is our initial model, which applies the QS-Attn on the last 2 layers. Model B involves all the five layers for QS-Attn. The best performance is indicated in \textbf{bold}.}
\end{table}

\begin{table}[t]
\centering
\resizebox{.9\columnwidth}{!}{
\begin{tabular}{@{}lccc@{}}
\toprule
\textbf{Method} & \textbf{Num of queries} & \textbf{FID$\downarrow$} & \textbf{SWD$\downarrow$} \\ \midrule
QS-Attn & \multirow{2}{*}{64} & 47.9 & 30.1 \\
CUT     &                      & 52.6 & 33.4 \\ \cdashline{1-4}
QS-Attn         & \multirow{2}{*}{128}       & \textbf{40.9}            & \textbf{28.5}            \\
CUT     &                      & 46.5 & 33.8 \\ \cdashline{1-4}
QS-Attn & \multirow{2}{*}{256} & 41.1 & 30.3 \\
CUT     &                      & 45.5 & 31.5 \\ \cdashline{1-4}
QS-Attn & \multirow{2}{*}{512} & 45.6 & 31.1 \\
CUT     &                      & 53.5 & 32.8 \\ \bottomrule
\end{tabular}
}
\caption{Ablation study for number of queries using in QS-Attn and CUT.}
\label{tab:number}
\end{table}

\section{Network Architectures}
In this section, we provide the network structure of our method, including the generator and discriminator. The detailed architectures of them are illustrated in \cref{sec:gen} and \cref{sec:dis}.

\subsection{Generator}\label{sec:gen}
Our generator $G$ consists of two down-sampling blocks, nine intermediate blocks, two up-sampling blocks and two convolution layers for input and output. We apply Instance Normalization (IN) \cite{ulyanov2016instance} and ReLU \cite{maas2013rectifier} in the generator, except for the output layer, which uses Tanh as the activation function. The network before the sixth residual block is regarded as the encoder $E$ and the rest is the decoder. We adopt the multi-layer feature extraction in CUT, which takes the features from five layers. including the input image, the first and second down-sampling blocks, and the first and fifth residual blocks. 

\subsection{Discriminator}\label{sec:dis}
We apply a PatchGAN \cite{son2017retinal} discriminator in our model, which contains three down-sampling blocks and two convolution layers. Leaky ReLU \cite{maas2013rectifier} is employed as the activation in the discriminator.

\begin{algorithm}[b]
\caption{{QS-Attn (Global)}}
\label{alg:global}
\begin{lstlisting}[language=python]
# H: height, W: width, C: dimension, 
# N: number of selected queries
# feat: input tensor (H, W, C)

feat_q = feat.flatten(0, 1)
feat_k = feat_q.permute(1, 0)
dots_global = torch.bmm(feat_q, feat_k)
attn_global = dots_global.softmax(dim=1)
prob = -torch.log(attn_global)
prob = torch.where(torch.isinf(prob), torch.full_like(prob, 0), prob)
entropy = torch.sum(torch.mul(attn_global, prob), dim=1)
_, index = torch.sort(entropy)
patch_id = index[:N]
attn_QS = attn_global[patch_id, :]
feat_out = torch.bmm(attn_QS, feat_q)
\end{lstlisting}
\end{algorithm}
\vspace{0.2cm}
		
\begin{algorithm}[b]
\caption{{QS-Attn (Local)}}
\label{alg:local}
\begin{lstlisting}[language=python]
# H: height, W: width, C: dimension
# N: number of selected queries
# k: window size
# feat: input tensor (H, W, C)

feat_local = F.unfold(feat, kernel_size=k, stride=1, padding=k//2)
L = feat_local.shape[1] # L = H * W
feat_k = feat_local.permute(1, 0).reshape(L, k * k, C)
feat_q = feat.reshape(L, C, 1)
dots_local = torch.bmm(feat_k, feat_q).squeeze(-1)
attn_local = dots_local.softmax(dim=1)
prob = -torch.log(attn_local)
prob = torch.where(torch.isinf(prob), torch.full_like(prob, 0), prob)
entropy = torch.sum(torch.mul(attn_local, prob), dim=1)
_, index = torch.sort(entropy)
patch_id = index[:N]
attn_QS = attn_local[patch_id, :].unsqueeze(1)
feat_v = feat_k[index, k * k, C]
feat_out = torch.bmm(attn_QS, feat_q).squeeze(1)
\end{lstlisting}
\vspace{-0.2cm}
\end{algorithm}

\begin{algorithm}[h]
\caption{{QS-Attn (Local+Global)}}
\label{alg:localglobal}
\begin{lstlisting}[language=python]
# H: height, W: width, C: dimension,
# N: number of selected queries
# k: window size
# feat: input tensor (H, W, C)

# Get patch_id from local attention.
feat_local = F.unfold(feat, kernel_size=k, stride=1, padding=k//2)
L = feat_local.shape[1] # L = H * W
feat_k = feat_local.permute(1, 0).reshape(L, k * k, C)
feat_q = feat.reshape(L, C, 1)
dots_local = torch.bmm(feat_k, feat_q).squeeze(-1)
attn_local = dots_local.softmax(dim=1)
prob = -torch.log(attn_local)
prob = torch.where(torch.isinf(prob), torch.full_like(prob, 0), prob)
entropy = torch.sum(torch.mul(attn_local, prob), dim=1)
_, index = torch.sort(entropy)
patch_id = index[:N]

# Select N rows in global attention matrix to route value.
feat_q_global = feat.flatten(0, 1)
feat_k_global = feat_q.permute(1, 0)
dots_global = torch.bmm(feat_q, feat_k)
attn_global = dots_global.softmax(dim=1)
attn_QS = attn_global[patch_id, :]
feat_out = torch.bmm(attn_QS, feat_q_global)
\end{lstlisting}
\vspace{-0.2cm}
\end{algorithm}

\section{Training details}\label{sec:training}
Under three settings of QS-Attn, \textit{Global}, \textit{Local}, and \textit{Local+Global}, we train the models for $400$ epochs with the batch size of $1$, the initial learning rate is set to $2 \times 10^{-4}$, and linearly decays to $0$ in the last $200$ epochs. All networks are optimized by the Adam optimizer \cite{kingma2014adam}, in which $\beta_{1}= 0.5$ and $\beta_{2}= 0.999$.

\section{More Results}
\begin{figure*}[t]
\centering
\includegraphics[width=1\textwidth]{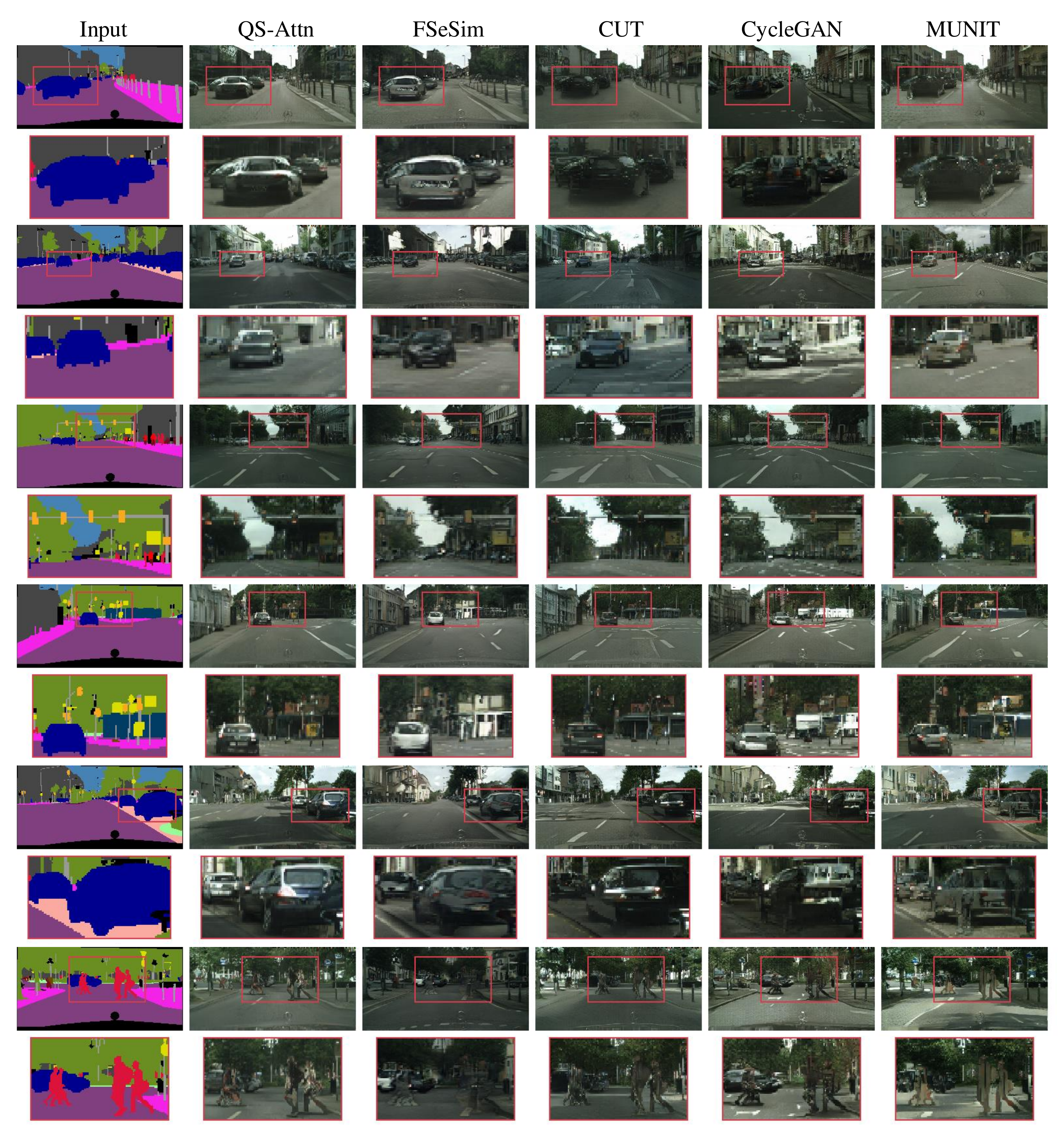}
\caption{Visual results on \textit{Cityscapes} compared with baselines. The leftmost column are the input source images. In the remaining columns, from left to right, are the translated results of our model, FSeSim, CUT, CycleGAN and MUNIT, respectively.}
\vspace{-0.4cm}
\label{fig:appcity}
\end{figure*}

\begin{figure*}[t]
\centering
\includegraphics[width=0.95\textwidth]{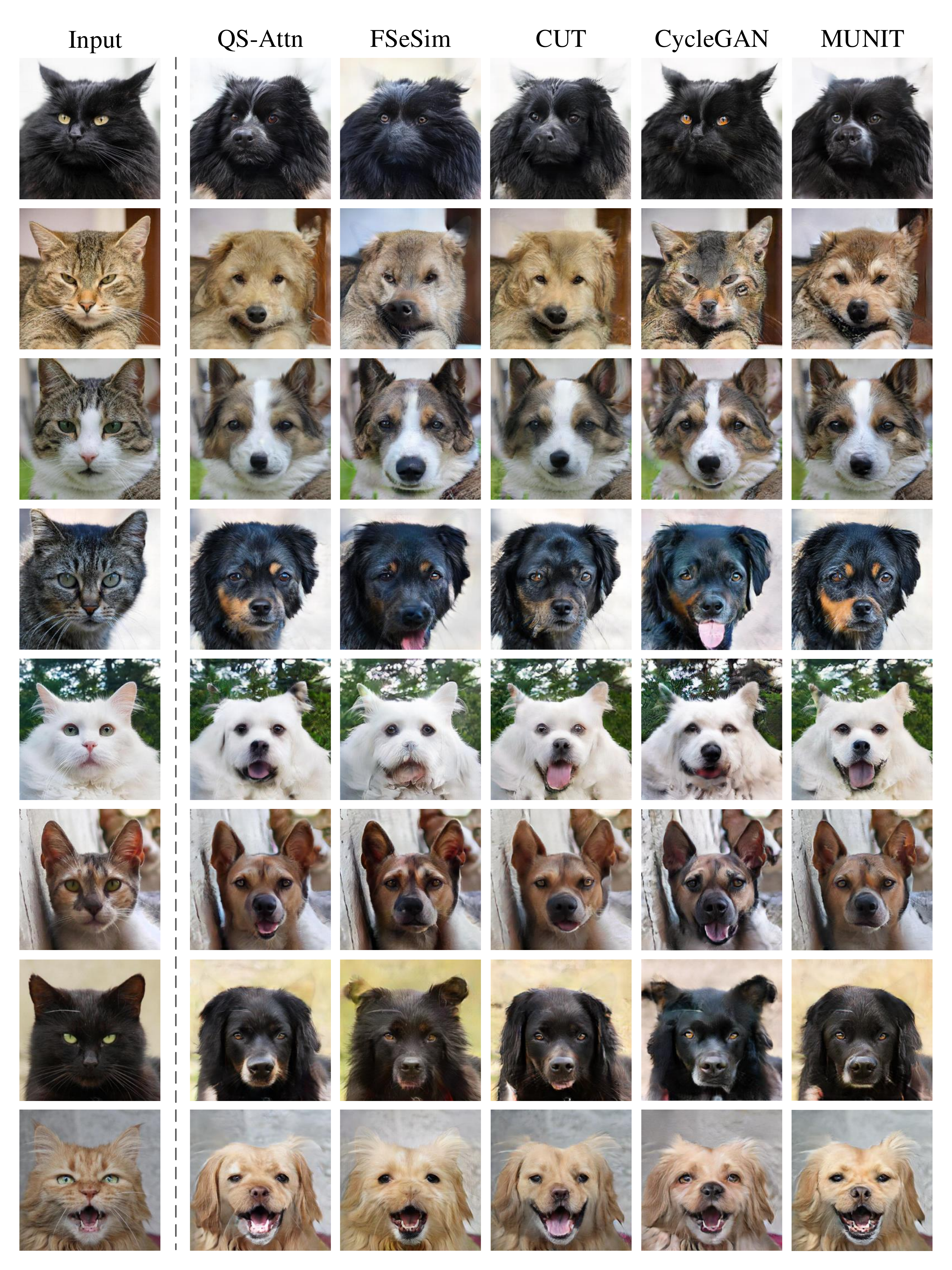}
\caption{Visual results on \textit{Cat $\rightarrow$ Dog} compared with baselines. The leftmost column are the input source images. In the remaining columns, from left to right, are the translated results of our model, FSeSim, CUT, CycleGAN and MUNIT, respectively.}
\label{fig:appcat}
\end{figure*}

\begin{figure*}[ht]
\centering
\includegraphics[width=0.95\textwidth]{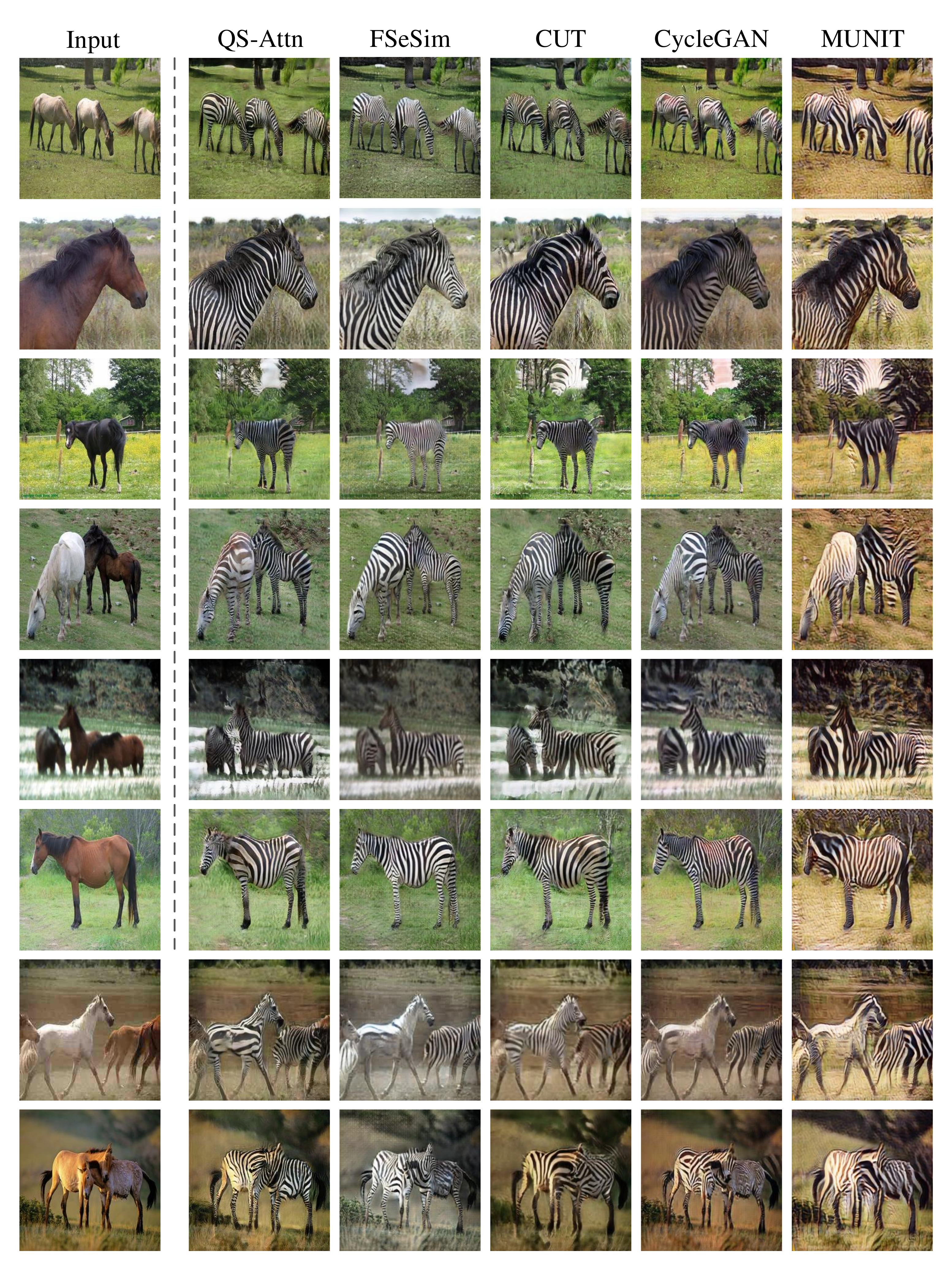}
\caption{Visual results on \textit{Horse $\rightarrow$ Zebra} compared with baselines. The leftmost column are the input source images. In the remaining columns, from left to right, are the translated results of our model, FSeSim, CUT, CycleGAN and MUNIT, respectively.}
\label{fig:apphorse}
\end{figure*}

Numerous synthesis results of our model on \textit{Cityscapes}, \textit{Cat $\rightarrow$ Dog} and \textit{Horse $\rightarrow$ Zebra} datasets are shown below. Besides, the visual results of baselines are also listed, including FSeSim \cite{zheng2021spatiallycorrelative}, CUT \cite{park2020contrastive}, CycleGAN \cite{zhu2017unpaired} and MUNIT \cite{huang2018multimodal}. \cref{fig:appcity}, \cref{fig:appcat}, \cref{fig:apphorse} illustrate the qualitative results on \textit{Cityscapes}, \textit{Cat $\rightarrow$ Dog} and \textit{Horse $\rightarrow$ Zebra} datasets, respectively.

\end{document}